# Deformable-Heatmap-Segmentation for Automobile Visual Perception

Hongyu Jin

*Abstract* — Semantic segmentation of road elements in 2D images is a crucial task in the recognition of some static objects such as lane lines and free space. In this paper, we propose. Deformable-Heatmap-Segmentation-Net(DHSNet),which extracts the objects' features with a U-shape end-to-end architecture along with a heatmap proposal. Deformable convolutions are also utilized in the proposed network. The DHSNet finely combines low-level feature maps with high-level ones by using upsampling operators as well as downsampling operators in a U-shape manner. Besides, DHSNet also aims to capture static objects of various shapes and scales. We also predict a proposal heatmap to detect the proposal points for more accurate target aiming in the network. The public dataset CityScapes is used to train and test our model. Comparisons between the proposed network and the UNet show that more detailed features are extracted by DHSNet, and its loss function also converges faster.

## I. INTRODUCTION

For the development of human being society, autonomous driving industry is one of the most worth engaging fields in the technical revolution. A safe and real-time running system of autonomous driving is promising for reducing traffic accidents and librating a large amount of labor force.

The semantic segmentation of road geometry and road scenes may indicate some critical information for autonomous driving automobiles, such as road element topology relationships, as well as potential traffic accidents between moving objects. But the dilemma of ODD[1] expanding, corner case solving and cost controlling has reached recently. The sheerly data driven problem solving methodology has proven nearly unaffordable. The call for better networks with comparatively simple structures arises.

Since nowadays witness a great blooming of computer vision tasks in the realm of vision perception[2] in auto-driving vehicles, demands for precise prediction of the surrounding environments grow. It is widely used in the modern automobile industry to apply 2D pixel-wise classification predictions to the revealing of road geometry and road scenes. The pixel-to-pixel inferencing conveys more detailed semantic information than object detection methods, with a thorough depiction of the topographical features.

However, some road geometries present extremely complicated structures and shapes, for instance, the poles and irregular obstacles. Therefore the semantic extraction methodology requires a sophisticated outline capture ability.

One of the major fashions in traditional pixel-wise prediction is CNN[3]. There is a huge shackle in the CNN method, that is, when the image is input to the network at the initial stage, because the convolution kernel of CNN is not large, the model can only use local information, always within a rectangular area(the traditional convolution is in the term of matrix with a fixed width and height), to understand the input image, which is inevitably a bit blind, thus affecting the distinguishability of the features extracted by the encoder at the end. This is a flaw that you can't escape just by using CNN. Of course, some plug-and-play modules based on the self-attention[4] mechanism can be inserted between the encoder and the decoder[5] to obtain the global context, so that the model understands the image from a global perspective and improve the features. However, if the model obtains the wrong features at the beginning due to a blind eye, there is a big question of whether it can be corrected by using the global context in the future.

Our proposed network DHSNet combines various useful structures in neural networks, such as deformable comvolution[6], heatmap proposal, multi-level feature

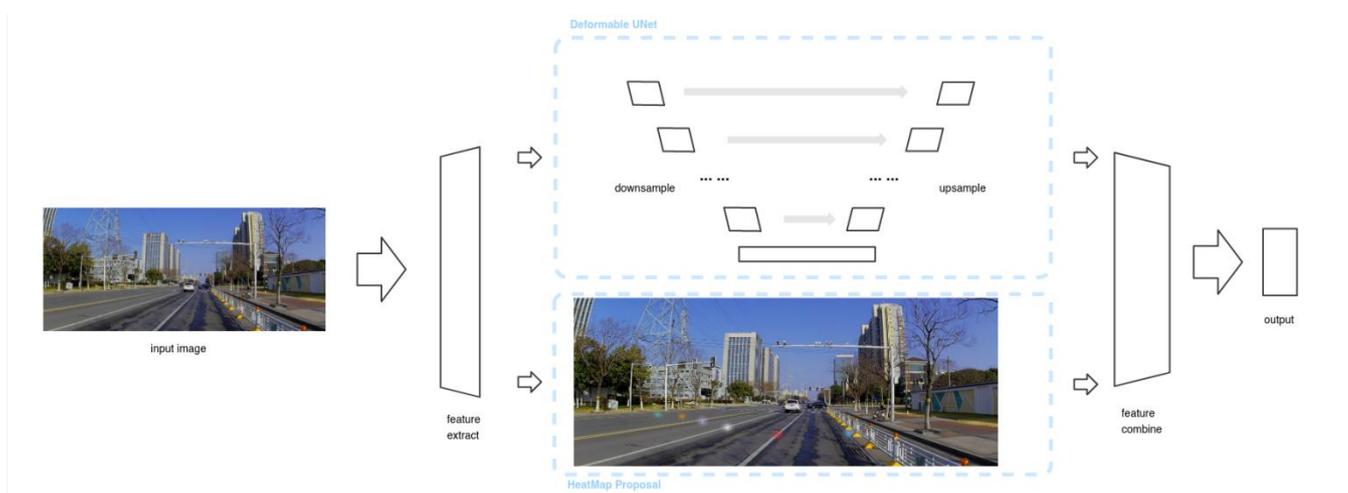

Fig. 1. DHSNet Architecture

combination[7], etc.

The deformable convolution is a structure which enables convolutions to have flexible shapes. It mainly consists of two proportions: regular convolution and a learnable deformable adjusting layer. The former one plays the role of weight storage and the approximate location for each weight unit. The latter provides an arbitrary shape depicting capability.

As for the auxiliary heatmap, the most "heated" areas are represented by a Gaussian kernel. The feature reaches the pick in the most center partition of kernel, and reduces its value as the Euclidean distance between candidate points and the center point grows. Therefore proposals are conducted. The proposal mechanism enables a precise prediction of the most focused areas of critical objects, the closest in-path vehicle for instance.

Our network also effectively captures multi-level neural network features by including feature fusion operation and multi-level feature refinement blocks. Feature fusion operators learn residual features and their combinations to fully exploit the complementary characteristics. Feature refinement blocks learn the combination of fused features from multiple levels to enable high-resolution prediction. Our network can efficiently train discriminative multi-level features end-to-end.

The above-mentioned proposed neural network structure shows the capability to solve some corner cases in the road-scene semantic segmentation task.

During the experimenting procedure, DHSNet stands out in 3-class semantic semantic tasks. First and foremost, it is able to draw vehicle, pedestrians and freespace areas in complicated shapes. In some cases, the cars in 2D images are divided by a pole or other agents on the road, the DHSNet can split the several parts very clearly, with less proportion missing due to the difficult shape extraction.There are also cases that some individual pedestrians are losing details in UNet semantic segmenting, whilst in DHSNet could maintain information reservation at a higher scale.

Moreover, the classification labels are inferences more accurately in DHSNet than in UNet. Compared with UNet outputs, which failed at times to configure background areas and mistaken them as vehicles or freespace. It is obvious that negligence in ways of these is unacceptable in safty-functional-driving tasks.

## II. RELATED WORK

### A. Deformable conv

The traditional convolution operation is to divide the feature map into parts of the same size as the convolution kernel, and then perform the convolution operation, and each part has a fixed position on the feature map. In this way, for objects with more complex deformations, the effect of using this convolution may not be very good.

For this case, the traditional approach is to enrich the data set, introduce more samples with complex deformations, use various data augmentations and tricks, and manually design some manual features and algorithms.

Dataset-based and data-augmented approaches are a bit "violent", usually slow to converge and require more complex network structures to cooperate, while algorithms based on manual features are a bit resource costing. The object detection[8] task and semantic segmentation task are complicated in real world, the deployment of traditional object detection and semantic segmentation faces corner cases especially when it comes to the irregular automobile and laneline tasks.

However, the structure of deformable convolution unit shows great potential in detecting objects in different shapes. Deformable Conv introduces an offset into the receptive field, and this offset is learnable, it can make the receptive field no longer a rigid square, but close to the actual shape of the object, so that the convolution area is always covered around the shape of the object, no matter how the object is deformed.

The potential capability of deformable convolution is revealed in the comparison between DETR and deformable DETR[9]. The deformable one gets performance improvements in the experiment . Deformable attention blocks gives DETR network a good chance to compete with CNN architecture.

Similar to Deformable attention block structure in deformable DETR, our network architecture adopts deformable convolution. Experiments also shows the ability of deformable convolution in general cases.

### B. Segmentation-based Methods

Deep learning has overcomes many computer vision tasks in industry, including automobile industry. The Deformable-Heatmap-Segmentation-Net (DHSNet) is designed for image semantic segmentation task on the latest vehicles' auto driving realm. The overall structure is a revolutionized from a classic semantic segmentation network UNet[10].

First, the UNet adopts series of downsampling[11] operations consisting of convolution and Max Pooling[12] , which named as the contracting path. The compression path consists of four blocks, each using three effective convolution and one Max Pooling downsampling, and the number of Feature Maps after each downsampling is multiplied by two, so there is a Feature Map size change shown in the figure. The final result is the down sampled Feature Map of the input image.

The expansive path is also composed of four blocks, each block starts by deconvolution to multiply the size of the Feature Map by two, while halving its number (the last layer is slightly different), and then merge with the feature map of the symmetric compression path, because the size of the compression path and the feature map of the extension path are not the same, UNet is by compressing the feature of the path The Map is normalized to feature maps of the same size as the extension path. The convolution operation of the extended path still uses the valid convolution operation.

Benefits from its effective network structure, UNet achieves applaudable performance in multiple semantic segmentation tasks. The advantage of this U-shaped structure comes from the combination of downsampling operation and deconvolution operation.

By this way of combination, UNet gathered together the features of deep and shallow layers. The deeper the network layer,  larger field of view the obtained feature map has. Shallow convolution focuses on texture features, and deep network focuses on more essential features, so deep shallow features are of equal significance;

Also, before U-Net, the edges information of the larger size feature map obtained by deconvolution are easy to be lack of information. In U-Net, However, each time the downsampling extracts the features, it will inevitably lose some edge features, and the lost features cannot be retrieved from the upsampling, so through the concat or adding operation of features, a retrieval of edge features is achieved.

*C. Heatmap-based Methods*

As shown in "Fig. 2", Heatmap-based Methods such as CenterNet detects the target as a point, that is, the center point of the target box represents the target. Predict the heatmap to represent the classification information. Each category has a heatmap, and on each heatmap, if there is a center point of the object target at a certain coordinate, a keypoint is generated at that coordinate. The keypoint will be denoted by a Gaussian circle, described as follow("Eq. (1)" ).

$$\hat{Y} \in [0, 1]^{\frac{W}{R} \times \frac{H}{R} \times C} \quad (1)$$

where R denotes the size of stride, C the number of keypoints type.

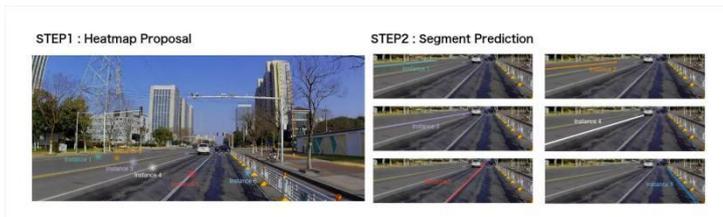

Fig. 2. Heatmap Proposal

( Assuming the input image to be $I \in R^{W \times H \times 3}$, with W being the width and H the height. ) A detection of object is revealed by $\hat{Y}_{x,y,c} = 1$ at the position.

During our training process, the ground truth label is in a form of heatmap. The annotation of the heatmap can be produced according to the bounding box

III. METHODS

The goal of our work is to build deep learning network architecture to do semantic segmentation in scenes of automobile driving. The semantic segmentation including laneline, vehicle, pedestrian, etc. We propose a new network named DHSNet for semantic segmentation task in automobile. The proposed approach is designed to integrate the advantages of both deformable unit, heatmap and U-Net architecture.

*A. Deformable Net basement*

The U-shape-like overall architecture is utilized as a fundamental base, for it is able to trade off the learning of deep layer features and shallow ones. Down sampling and up sampling are on the early and later part of the whole network. The following figure depicts the U-shape structure.

On the basis of U-shape structure, we exchange the original convolutional blocks with deformable convolutional blocks in multiple location during the encoder process, which is also called, the down sampling operation. The up sampling half, however, keeps the origin structure of UNet.

Furthermore, a heatmap scheme is trained at the same time for classification refinement in the main stream semantic segmentation procedure. The classification information in respective zone on the input image will be predicted by heatmap, feeding to the segmentation process afterwards. This plays a role as similar as NMS in object detection tasks.

The heatmap scheme consists of a Resnet18[13], a convolution layer ( 3*3 conv kernel shape, strides[1,1], padding[1,1,1,1]), a Relu as activation layer and another convolution layer ( 3*3 conv kernel shape, strides[1,1], padding[1,1,1,1]). The ground truth of the heatmap stem is in a form of a mask map with several Guassian Kernels on it. Each Guassian Kernel indicates the classification type and the location of objects that are tended to be segmented by main stream semantic segmentation task.

- **U-shape basic architecture.**

Similar to UNet, the DHSNet combines convolutional Encoder and the deconvolutional Decoder to extract semantic features step by step. With 4 Encoder blocks, the network's middle output conveys higher-level information.

At the same time, each separate Encoder block transfer the feature to their correspondingly homogeneous Decoder blocks which hold same-dimensions as the Encoder ones. The gap between early-stage down sampling and later-stage up sampling will merge and finely concate to calculate the final output. For instance, the edge information always get lost after a long way of network calculation, but the bridge between Encoder and Decoder deals subtly with this edge feature vanishing problem.

- **Deformable Convolutional Blocks.**

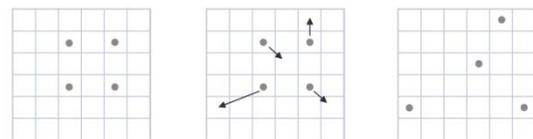

Fig. 3. How Deformable Convolutions works

UNet performs extraordinarily well in many realms, especially in medical use. Nevertheless, little difficulties are identified whilst making efforts to apply this network structure for segmentation tasks in automobile domains. Obstacles have various 2D shapes, pedestrian and truck are hard to share outlines with high likelihood. No need to mention the Lane Lines with a long and slim feature. The merely usage of UNet lacks flexibility when facing real world objects of vast categories. Thus, we utilize deformable convolution in each Decoder block.

Deformable Convolutional Block has the ability of depicting various shaping. It adds an offset to each point on the receptive field, and the size of the offset is learned by learning, and the receptive field is no longer a square, but matches the actual shape of the object. The advantage of this is that no matter how the object is deformed, the area of convolution is always covered around the shape of the object.

The traditional convolutional structure can be defined as "Eq. (2)", where each point of the output feature map corresponds to the center point of the convolution kernel and pn is each offset of p0 within the convolution kernel range.

$$y(p_0) = \sum_{p_n \in R} w(p_n) \cdot x(p_0 + p_n) \quad (2)$$

Deformable convolution, on the other hand, introduces an offset for each point based on "Eq. (2)" above, which is generated from the input feature map and another convolution, usually a decimal.

$$y(p_0) = \sum_{p_n \in R} w(p_n) \cdot x(p_0 + p_n + \Delta p_n) \quad (3)$$

Since the position after adding the offset is not an integer and does not correspond to the actual pixels on the feature map, it is necessary to use interpolation to obtain the offset pixel value, which can usually be bilinear interpolation, which is expressed by the following formula "Eq. (4)":

$$\begin{aligned} x(p) &= \sum_q G(q, p) \cdot x(q) \\ &= \sum_q g(q_x, p_x) \cdot g(q_y, p_y) \cdot x(q) \\ &= \sum_q \max(0, 1 - |q_x - p_x|) \cdot \max(0, 1 - |q_y - p_y|) \cdot x(q) \end{aligned} \quad (4)$$

The meaning of the above formula is to set the pixel value of the interpolation point position to the weighted sum of its four field pixels, the field four points are the nearest pixel that actually exists on the feature map, and the weight of each point is set according to its distance from the horizontal and vertical coordinates of the interpolation point, and the max(0, 1-...) of the last line of the formula restricts the distance between the interpolation point and the field point by no more than one pixel.

- Auxiliary Heatmap.

Use Heatmap to locate the object's center, thus help with the total output. In this way, no nms steps are needed in the whole inference pipeline.

In non-deep learning fields, heat mapping refers to simply aggregating large amounts of data and elegantly representing it using a progressive color ramp to visualize how dense or frequent spatial data is. In deep learning studies, heatmaps help to understand which part of an image allows the neural network to make the final classification decision. There are two ways to generate a heat map, one is a Gaussian heat map and the other is a class activation heat map generated by Grad-CAM[14].

In this paper we adopt Gaussian heat map. Observing the Gaussian kernel formula("Eq. (1)" ) used to map the heatmap, if we take (x,y) to be the step block position enumerated in the image to be inspected, and $(p_x, p_y)$ is the coordinates corresponding to the GT key in the image. The annotated key points of objects will be distributed in the form of "Eq. (5)" in feature map:

$$Y_{xyz} = exp(-\frac{(x - \tilde{p}_x)^2 + (y - \tilde{p}_y)^2}{2\sigma_p^2}) \quad (5)$$

When the position of the enumeration block and the ground trueth key point coordinates are close to coincide, the Gaussian kernel output value is close to 1; When the enumeration block position and the ground truth key are very different, the Gaussian kernel output value is close to 0.

As for the loss function for this proposal partition, is as the following function:

if $\hat{P}_{xy}=1$,

$$L_{point} = -\frac{1}{N_p} \sum_{xy} (1 - \hat{P}_{xy})^\alpha \log(\hat{P}_{xy}) \quad (6)$$

otherwise,

$$L_{point} = -\frac{1}{N_p} \sum_{xy} (1 - P_{xy})^\beta (\hat{P}_{xy})^\alpha \log(1 - \hat{P}_{xy}) \quad (7)$$

Although heatmap shows great potential on figuring out centers and classifications of objects, it lacks the capability of drawing out the edges. Whereas, the segmentation partition holds its merit of an accurate shape depicting ability and weakness in classification prediction.

The DHSNet adapts a combination of segmentation and heatmap. Our network designment leverages both methods, promoting the benefits from the deformable segmentation network method and heatmap method with improved perception result while overcoming their limitations.

The heatmap result plays the role of proposal point, the points are located on the center of main objects. To constraint the proposal heatmap, we adopt focal loss[15] following CenterNet [16].

IV. EXPERIMENTS

*A. Experiment Setting*

*1) Datasets*

We utilize CityScapes[17] dataset as the training and testing dataset. Cityscapes has 5,000 images of driving scenes in urban environments (2975train, 500 val, 1525test). It has dense pixel annotation (97% coverage) for 19 categories, 8 of which have instance-level segmentation. The Cityscapes dataset, or Cityscape dataset, is a new large-scale dataset that contains a set of different stereoscopic video sequences recorded in street scenes in 50 different cities.

The cityscapes dataset is centered on the semantic understanding picture dataset of urban street scenes, which contains a variety of stereoscopic video sequences recorded in street scenes from 50 different cities, including 5,000 frames of high-quality pixel-level annotation in addition to 20,000 weakly annotated frames. As a result, datasets are orders of magnitude larger than previous datasets. The Cityscapes dataset has two sets of evaluation criteria, the former providing 5,000 finely labeled images, and the latter providing 5,000 finely labeled images plus 20,000 coarse-labeled images.

*2) Evaluation Metrics*

We evaluated our model using Accuracy (ACC) "Eq. (8)" as metrics to statistically evaluate the performance of models. ACC is a metric for measuring the ratio between the correctly classified pixels and the total pixels in the dataset. The metrics have the forms as following:

$$ACC = \frac{\sum_{i=0}^{n}(TP+TN)}{\sum_{i=0}^{n}(TP+FP+TN+FN)} \quad (8)$$

Where TP represents the number of the true positive samples; TN stands for the number of the true negative samples; FP means the number of the false positive samples; FN means number of the false negative samples. Meanwhile, n indicates the total classification categories in the whole system.

*3) Device and other basic settings*

The experiments' trainning process ran on 8 A100 cards, with a batch size of 16. The evaluation ran on 1 A100, cardwith a batch size of 1.

## B. Results

Our experiments evidently show the effectiveness of DHSNet's structure by comparing it with the UNet[2]. The most representative experiment results will be exhibited below, including the totally calculated metrics results, the typical comparative study, and the loss stats.

*1) Metrics results*

The metrics results on the cityscapes dataset are shown in the "Table. 1". It is obvious that the capability of pixel-wise semantic segmentation of DHSNet is remarkably outstanding, compared with UNet, a widely utilized semantic segmentation neural network.

TABLE I.  METRICS RESULTS

| Model | Dataset | ACC |
|---|---|---|
| UNet | Cityscapes | 0.93568 |
| DHSNet | Cityscapes | 0.94831 |

*2) Detailed visualization results*

The visualization results on the dataset are shown below. The results show that our method can cope with complex line topologies and irregular shaping objects. Even for the cases of dense objects and fork objects, our method can also successfully discriminate the instances.

Firstly, the somehow partly covered vehicle objects are better segmented in DHSNet than in UNet. Take "Fig. 4" and "Fig. 5" as examples, in "Fig. 4", the vehicle is blocked by a traffic sign on the front. In DHSNet, the vehicle object as background and the traffic sign onject as front-ground both shares a more accurate shape depiction. While in UNet, the same 2D image produces a thinner shape of traffic sign, where detailed semantic information misses.

in "Fig. 5", a pole is in the middle between the ego car and the vehicles on opposite lanes. The UNet ignores most pixels of the slim and long pole. However, DHSNet conduct the rough shape of pole. Thus, DHSNet's ability of extracting lines features and other slim-shape object is testified and proved.

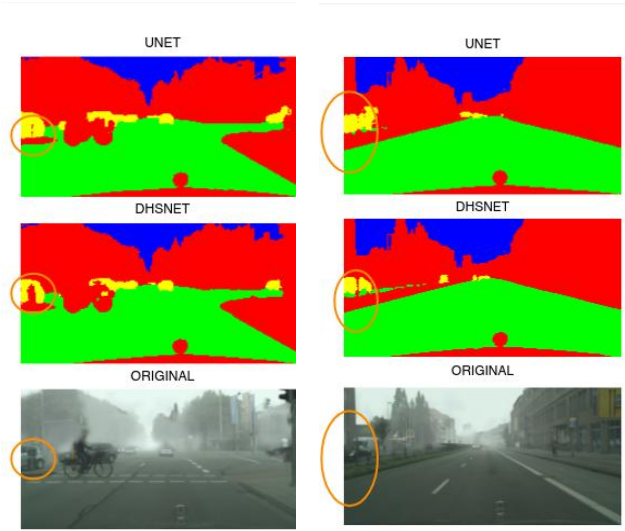

Fig. 4. (Left) Vehicle split by a traffic sign

Fig. 5. (Right) Vehicle split by a pole

In "Fig. 6", vehicle objects with fuzzy features of shape and outlines vanishes after the network's calculation in UNet.

Analyze integratedly with the afore mentioned "Fig. 4" and "Fig. 5",it is not only a typical phenomenon in the picture in this paper, but also a generally regular pattern that the failure of capturing features of split or blocked vehicle as well as the somewhat indistinct vehicles(maybe vehicles moving too fast) are more often occured close to the left and right edge of 2D images.

Apart from semantic segmantation in vehicle objects, the pedestrian task is more challenging. The orignal picture in "Fig7" has 2 human being and a small cart between them. The DHSNet extracts more fetures of those three objects. The shapes of leges of pedestrians are more natrual, closer to reality. The cart's wheels are clearer, too. Nevertheless, in the experiment of this paper, the overall shapes of pedestrains are not satisfying. more experiments should be done afterwards.

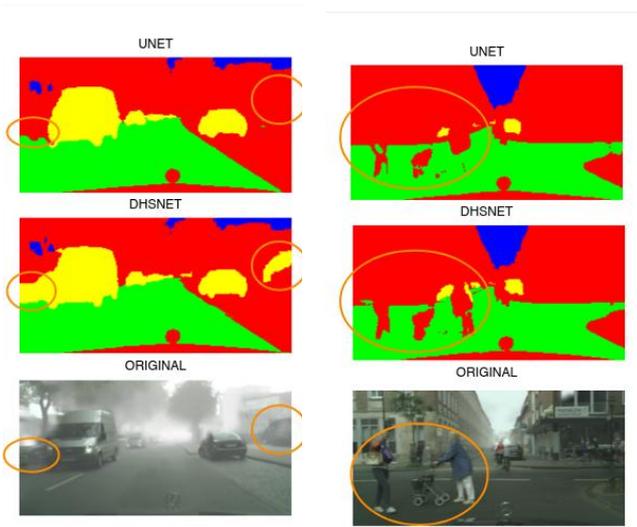

Fig. 6. (Left) Fuzzy vehicles

Fig. 7. (Right) Pedestrians

In "Fig. 8", a truncation vehicle only has a tailstock of it.

UNet depicts most proportion of this car, but it is worth mentioning that the entire tailstock is broken into peices.

The lack of coherence between the three seperated vehicle partitions comparitively illustrates that linking the seperated partition of one object is also a strength of DHSNet.

The omission of details on the left hand side in the UNet part in "Fig. 9" is shown. Which illustrates that not only does the DHSNet excel at percepting vehicles, pedestrians and freespaces, but also out stands in other objects in surrounding environments of road scenes.

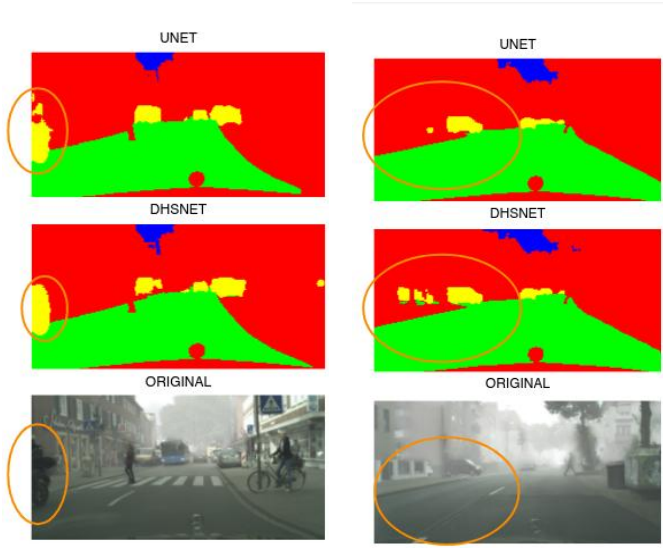

Fig. 8. (Left) Missing car

Fig. 9. (Right) Missing other objects

*3) Loss record*

Since the loss function also converges faster. reveals the fact that the deformable convolution depicts irregular shaping better than the traditional rectangle convolutions.

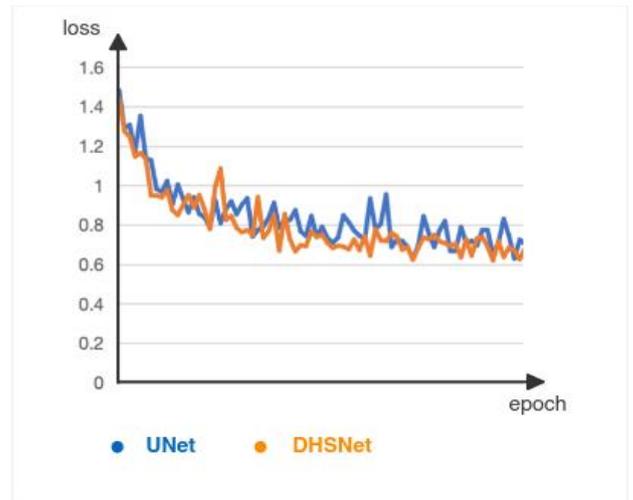

Fig. 10. Loss record conparison

*C. Time consume*

Though the DHSNet increases the accuracy in road-scene-based segmentation tasks, meanwhile, it is a inevitable shortcoming of this network structure to be more time cosuming than the easier-structured UNet.

TABLE II. INERENCE TIME

| Model | Device | avg timing (second) |
|---|---|---|
| UNet | 1 A100 | 0.0233 |
| DHSNet | 1 A100 | 0.0724 |
| UNet+1Deform Convalution | 1 A100 | 0.0250 |
| UNet+2Deform Convalution | 1 A100 | 0.0314 |

## V. CONCLUSIONS

The DHSNet architecture achieves very good performance on Cityscapes dataset's road-scene-based segmentation applications on accuracy. The inderencing devices are remain unchanged betwwen different models. More inprovements and tactics should be applied to make the model inferencing of DHSNet more real-timing.